\documentclass[a4paper,12pt,final]{article}

\pdfoutput=1
\usepackage{showkeys}
\usepackage[utf8]{inputenc}
\usepackage[T1]{fontenc}
\usepackage{wrapfig}
\usepackage{cite} 
\usepackage{amsmath}
\usepackage{lmodern}
\usepackage{units}
\usepackage{icomma}
\usepackage{amssymb}
\usepackage{floatrow}
\usepackage[font={small, it}]{caption}
\usepackage{color}
\usepackage{caption}
\usepackage{alltt}
\usepackage{subcaption}
\usepackage{hyperref}
\usepackage{cleveref}
\usepackage{multirow}
\usepackage{graphicx}
\usepackage{bigints}
\usepackage{epstopdf}
\usepackage{bbm}
\usepackage{marvosym}
\usepackage[version=3]{mhchem}
\usepackage[ruled]{algorithm2e}
\usepackage[english]{babel}
\newcommand{\argmin}{\mathop{\mathrm{argmin}}}

\usepackage{hyperref}
\usepackage{float}
\usepackage{hhline}
\usepackage{fancyhdr}
\usepackage[margin=0.7in]{geometry}
\usepackage{geometry}
\usepackage{pdflscape}
\usepackage{multirow}
\usepackage{multicol}
\usepackage{listings}
\usepackage{tikz}
\usetikzlibrary{shapes.geometric, arrows}
\tikzstyle{node} = [rectangle, rounded corners, minimum width=3cm, minimum height=1cm,text centered, draw=black, fill=gray!20]
\tikzstyle{g_node} = [circle, rounded corners, minimum width=1cm, minimum height=1cm, text centered, draw=black, fill=gray!20]
\tikzstyle{in_node} = [circle, rounded corners, minimum width=1cm, minimum height=1cm, text centered, draw=black]
\tikzstyle{arrow} = [thick, ->, >=stealth]
\tikzstyle{logical} = [rectangle, rounded corners, minimum width=3cm, minimum height=1cm,text centered, draw=black]
\tikzstyle{result} = [diamond, minimum width=3cm, minimum height=1cm,text centered, draw=black, fill=gray!20]
\usepackage{color}
\definecolor{codegreen}{rgb}{0,0.6,0}
\definecolor{codegray}{rgb}{0.5,0.5,0.5}
\definecolor{codeblue}{rgb}{0,0.5,1}
\definecolor{codepurple}{rgb}{0.54,0.17,0.89}
\definecolor{codeorange}{rgb}{1,0.58,0}
\definecolor{backcolour}{rgb}{1,0.98,0.94}

\newtheorem{definition}{Definition}
\newtheorem{property}{Property}

\usepackage{graphicx}
\usepackage{epsfig}
\usepackage{amsmath,amsfonts}
\usepackage{url}
\usepackage{pslatex}

\def\NVcom#1{{{}}\\}

\usepackage[T1]{fontenc} 

\def\NVcom#1{{{}}\\}

\makeatletter
\renewcommand{\thetable}{\@arabic\c@table}
\@addtoreset{table}{section}
\makeatother
\title{Latent space conditioning for improved classification and anomaly detection
}

\author{\bf Erik Norlander and Alexandros Sopasakis\\
Lund University}

\begin{document}

\thispagestyle{empty} 
\maketitle

\begin{abstract}
	We propose a variational autoencoder to perform improved pre-processing for clustering and anomaly detection on data with a given class label. However, in this case, anomalies are not known or labeled. We call our method the Conditional Latent Space Variational Autonencoder since it separates the latent space by conditioning on information within the data. The method fits one prior distribution to each class in the dataset, effectively expanding the prior distribution to include a Gaussian mixture model.
	Our approach is compared against the capabilities of a typical variational autoencoder by measuring their V-score during cluster formation with respect to the $k$-means and EM algorithms.
	For anomaly detection, we use a new metric composed of the mass-volume and excess-mass curves which can work in an unsupervised setting. We compare the results between established methods such as as isolation forest, local outlier factor and one-class support vector machine. 
\end{abstract}


\section{Introduction\label{intro}}  
Most anomaly detection and clustering methods will perform well on good data. Real life applications however do not follow any guidelines. Actual data is typically incomplete, faulty, missing, unbalanced or consist of outliers. Data can have a large amount of mixed categorical and numerical features making it difficult to ascertain their relational importance. Many times there exist a size imbalance between categories. Thus a machine learning model could bias towards the larger classes in the data set. Furthermore overlap can make it difficult to identify the class each of the data points belongs to. In this case, anomaly detection can be challenging since an anomalous data point for one class may be a normal point for another. Last but not least data could be mislabeled which in turn can lead to severe modeling errors.

Clustering is a classic machine learning problem that is well studied but it is also heavily data dependent \cite{ITISESop}. The success of the clustering often comes down to feature engineering or otherwise pre-processing the data in some way if it's not already nicely placed in some euclidean space. 

Anomaly detection is one of the most important problems in manufacturing \cite{Mart}, cyber security \cite{Schubert}, medical imaging \cite{Zenati}, fraud detection and several others. At the same time anomaly detection is a difficult problem since it is heavily dependent on both data quantity and quality. The problem becomes even more complicated when considering an \textit{unsupervised} setting such as we do here.

In this paper, we consider data where the class label is known but we do not know which points are anomalous and which are normal. Consequently, it is important to specify what we mean by \textit{anomaly}. In this paper, we will consider anomalies to be data points which occur in low probability regions of the dataset. Moreover, the proposed framework will be able to work with a broad category of data in an unsupervised manner. It will also allow us to use established methods of clustering and anomaly detection for comparison purposes.  


We propose to use a variation of the VAE \cite{VAE} for this purpose. Instead of using the generative aspects of the VAE, we will perform our analysis in \textit{latent space}. This is a reduced dimensionality space which forces the data to be close to some prior distribution. The latent space will also retain relational information of features. We propose to name this algorithm \textit{Conditional Latent Space Variational Autoencoder} or CL-VAE for short. 

We begin in Section \ref{methodology} with a brief overview of the clustering and anomaly detection metrics which we apply during this study. In Section \ref{sec:vae} we present the theory behind VAE's and in particular how exactly we condition our VAE to fit multiple Gaussians and categorize the latent space based on the data. The choice of loss function is critical in forming the clusters. In Section \ref{sec:cvae} we provide the theoretical background on the Kullback-Leibler (KL) divergence used for regularization as well as the reconstruction piece of the loss function. We present the unsupervised case of anomaly detection in Section \ref{sec:anomaly} where we also discuss the proposed EM-MV measure to evaluate its performance. We end with a number of remarks and an overview in Section \ref{discussion}.

\section{Approach and state of the art\label{methodology}}

There are a number of approaches to anomaly detection using generative models; many, using the reconstruction error as a measure of "degree" of abnormality \cite{bologna, NUS}. However in this paper we implement a VAE as a pre-processing algorithm for use in common anomaly detection algorithms. In order to do so effectively, we introduce the CL-VAE which shapes the latent space by specifying which distribution each data point belongs to, making it easier to work with. 

The idea of using a Gaussian Mixture Model (GMM) as prior for training a VAE is not a new idea \cite{RuiShu, GMVAE}. However, in the case of GMVAE \cite{GMVAE}, there is no conditioning on a pre-defined label. Rather, it tries to estimate the prior distribution with Monte Carlo methods. A similar algorithm was cleverly described by J. Su \cite{one-step} but it includes a classifier that we do not really need here since we know which class to assign to which Gaussian. This is also related to the DEC-algorithm by J. Xie et al. \cite{DEC} and the VaDE-algorithm by Z. Jiang et al. \cite{VaDE} with a  built-in unsupervised cluster assignment which is not something that we attempt here. Our algorithm is different from the CVAE (Conditional Variational Autoencoder) in \cite{CVAE, Tut} because we're conditioning the latent space itself on the class label by selecting the appropriate Gaussian, not assuming we can map all classes to the same Gaussian as CVAE does. Instead the CVAE conditions the encoder and decoder, keeping the one-gaussian assumption in the latent space which we avoid.

Two different datasets will be used in this paper in order to illustrate different properties of the proposed methodology. First, the standard MNIST dataset will be used to present  clustering and anomaly detection capabilities in order to compare with other publications and results on that dataset.  The second dataset we use consists of actual trades made by traders at a Swedish investment bank. Thus this data set is typical of the many problems a real dataset can have: unbalanced classes, mislabeling, missing data etc. This dataset was also used in the thesis that this paper is based on \cite{erno} and is included to diversify the discussion on anomaly detection as well as to illustrate a real application of the proposed approach.

We will first train and produce latent spaces from the two different autoencoders: VAE and CL-VAE. We then  compare how well we can perform clustering and anomaly detection on each. For the sake of visualization we have chosen to present the results in 2 latent dimensions. We note however that performance could increase if we were to use more dimensions in latent space as it would include more information. Arguably two of the most standard clustering algorithms, which we also use in this paper, are the $k$-means \cite{k-means} and EM \cite{em} algorithms. Similarly, for anomaly detection the most frequently used algorithms in this area are Isolation Forest \cite{isolation_forest}, LOF \cite{LOF} and OCSVM \cite{OCSVM}.

\subsection{Detection methodology\label{metrics}}

A typical issue for most clustering algorithms is how to evaluate their results. This is because, while clustering itself is an unsupervised task, in some cases we can check the clusters that do form against some ground truth. In this case, that would be class label.

The V-score or validity score is one such metric capable of measuring this kind of clustering result. It is an entropy \eqref{def:entropy} based approach and works by measuring two interdependent characteristics in clusters: \textit{completeness} and \textit{homogeneity} \cite{v-score}. Homogeneity measures whether every member of a given cluster belongs to a single class. This is done as follows,
$$
h = \label{def:homo}
\begin{cases}
1, & \mbox{if } \quad H(C,K) = 0,  \\
1 - \frac{H(C|K)}{H(C)}, & \mbox{otherwise.}
\end{cases}
$$
Note that the case with cross entropy of $H(C|K) = 0$ denotes a totally homogeneous cluster. Completeness on the other hand measures whether all known members of a class are assigned to the same cluster. We define completeness via,
$$
c = \label{def:comp}
\begin{cases}
1, & \mbox{if } \quad H(K,C) = 0,  \\
1 - \frac{H(K|C)}{H(K)}, & \mbox{otherwise.}
\end{cases}
$$



Note that $0\leq c \leq 1$. In practice however we normalize the conditional entropy by $H(C)$ in order to remove any class size dependencies. The validation score or V-score is defined to be the weighted mean between $c$ and $h$,
\[
V_\beta = \frac{(1+\beta) h c}{\beta h + c}, \label{def:V-score}
\]
where $\beta$ is a weight favoring completeness if $\beta>1$ or homogeneity if $\beta <1$. Readers might recognize how the completeness and homogeneity defined above encompass metrics such as inertia and the Dunn index \cite{Dunn} which have been used in the past to measure intra-clustering and inter-clustering distances. 


\subsection{Anomaly detection  \label{F1metrics}}
Confirming anomalies which are not already labeled is not a trivial task. In this unsupervised setting we can not use Precision-Recall or ROC simply because there's no way to check our results against a ground truth. Therefore, the excess mass-mass volume (EM-MV) method can be a great tool at indicating anomalies as it has been found to agree with its supervised counterparts \cite{EM-MV}. 

We assume that anomalies occur on the tail ends of a probability distribution. So our goal is to estimate the density level curves of that distribution. The method works on level sets for which a function $f$ equals a given constant $c$, $L_c(f) = \{ (x_1, \dots, x_n) \mid f(x_1,\dots, x_n) = c\}$. In our case the function $f$ is the probability density which we estimated by our CL-VAE in latent space. The degree of abnormality or anomaly for the EM-MV method is given by a scoring function $s: \mathbb{R}^d \xrightarrow{} \mathbb{R}_+$ for data in $\mathbb{R}^d$. The method relies on the mass volume (MV$_s$) and excess mass (EM$_s$) curves of $s$ as follows,
\begin{align}
        MV_s(\alpha) = \inf_{u \geq 0} \quad \text{Leb}(s \geq u) \text{s.t.} \quad \mathbb{P}(s(X) \geq u) \geq \alpha, 
        \label{def:MV}
\end{align}
\begin{align}{}
        EM_s(t) = \sup_{u \geq 0} \quad \{\mathbb{P}(s(X) \geq u) - t\text{Leb}(x \geq u)\},
        \label{def:EM}
\end{align}
where $t>0, \alpha \in (0,1)$ and $\mathbb{P}(s\geq t)=\frac{1}{n}\sum_{i=1}^n \mathbbm{1}_{s(X_i)\geq t}$. So now we can evaluate our method by computing the distance between the level sets of $f$ and $s$ for excess mass and mass volume as follows: $||EM_s - EM_f||_{L^1(I)}$ and $||MV_s - MV_f||_{L^1(J)}$ \cite{EM-MV}. Practically we let $I=[0.9,0.999]$ and $J=[0, EM_s^{-1}(0.9)]$ where $EM^{-1}_s (0.9) =  \inf \{t \geq 0, EM_s(t) \leq 0.9\}$ as shown in \cite{EM-MV}. One technical issue involves the computation of the Lebesgue terms above which can be resolved via Monte Carlo estimation.

The measure is then based on the area under the EM$_s$ and MV$_s$ curves. This area should be \textit{maximized} for EM$_s$ and \textit{minimized} for MV$_s$.

\newpage
\section{Creating Generative Models\label{sec:vae}}
As mentioned earlier, Variational Autoencoders (VAEs) are generative models introduced in 2014 by Kingma and Welling \cite{VAE}. The purpose of such a model is density estimation. Meaning, that a VAE can estimate any underlying distribution in a dataset and start to generate new samples from that dataset. 

\begin{wrapfigure}[]{r}{0\textwidth}
	\centering
	\resizebox{.25\linewidth}{9cm}{%
	\begin{tikzpicture}[node distance=1.5cm]
	\node (output) [in_node] {$\hat{x}$};
	\node (decoder) [node, above of=output] {Decoder};
	\node (z) [g_node, above of=decoder,] {$z$};
	\node (mu) [in_node, above of=z, xshift=20] {$\mu$};
	\node (sigma) [in_node, above of=z, xshift=-20] {$\Sigma$};
	\node (epsilon) [g_node, left of=sigma] {$\epsilon$};
	\node (encoder) [node, above of=z, yshift=50] {Encoder};
	\node (input) [in_node, above of=encoder] {$x$};
	\draw [arrow] (input) -- (encoder);
	\draw [arrow] (encoder) -- (mu);
	\draw [arrow] (encoder) -- (sigma);
	\draw [arrow] (epsilon) -- (sigma);
	\draw [arrow] (mu) -- (z);
	\draw [arrow] (sigma) -- (z);
	\draw [arrow] (z) -- (decoder);
	\draw [arrow] (decoder) -- (output);
	\end{tikzpicture}
	}
	\caption{Diagram of a VAE structure. The encoder neural network maps the input $x$ to the latent space described by the Gaussian distribution $z$. Then the Decoder neural network creates a reconstruction of the input in $\hat{x}$.}
	\label{fig:VAE}
	
\end{wrapfigure}
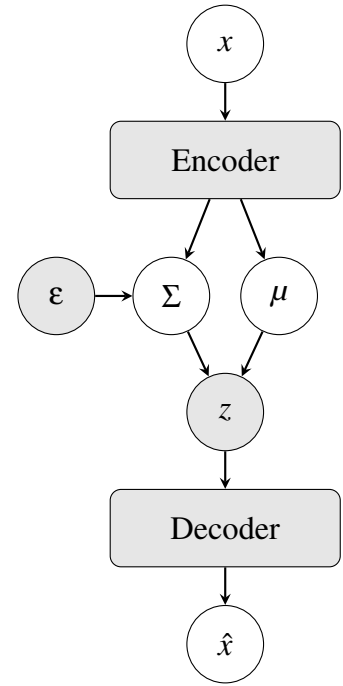
The VAE consists of a encoder-decoder pair of neural networks, with a stochastic latent layer in the middle as depicted in Figure \ref{fig:VAE}. The encoder which we express here with $p_{\theta}(z|x)$ processes the input $x$ and produces the parameters of a Gaussian distribution represented by $z$ in the latent layer as shown in Figure \ref{fig:VAE}. The decoder which we denote by $p_{\theta}(\hat{x}|z)$ uses that Gaussian distribution $z$ in latent space as input to produce the parameters to an approximation of the probability distribution describing the original data. The weights and biases of the encoder and decoder networks are represented in $\theta$. 

VAEs allow us to effectively compress information within the data through dimensionality reduction occurring in the latent space. Since the decoder goes from a smaller to a larger space information is lost. We measure this loss by the reconstruction loss which is the log-likelihood of $p_{\theta}(\hat{x}|z)$ which will give us a loss function to minimize in training. This shows also how effective the encoder was in compressing information within $z$ from the original data $x$. Furthermore VAEs can also be used with categorical data as well as non-linear type transformations in contrast to PCA type methods \cite{PCA}.

Note that the VAE architecture is very similar to that of an autoencoder shown in Figure \ref{fig:my_label} in the Appendix except for the latent space which is sampled by estimating $\mu$ and $\Sigma$. This is done by mapping the mean and variance of these points with the stochastic layer described by $\epsilon$, $\mu$ and $\Sigma$. We also note that every point in the latent space of a VAE is forced to have feasible features since it is assigned a variance. A prime advantage of this approach is that VAEs, unlike regular autoencoders, map values to the latent space by retaining relational information between them so that they remain meaningful for analysis purposes \cite{Chollet}.

Another type of generative model is a GAN \cite{GAN} that solves the same problem with a game-theoretic approach. The VAE on the other hand is derived from a purely statistical point of view and the fact that you can use if for generation is almost just a bonus \cite{TrainVAE}. Because of the statistical interpretability and ease of manipulation, we chose to use a VAE for this task.

\subsection{The optimization function\label{sec:loss}}

The choice of loss function during any optimization or neural network problem is critical and has direct implications in success or failure of the given problem. To create a proper loss function however requires knowledge of $p_{\theta}(x)$ in order to compute the $\log p_{\theta}(\hat{x}|z)$ which is intractable (see Appendix B). To counter this problem we define a lower bound of the likelihood and optimize that instead.

Assuming that $z$ can be estimated using the distribution $q_\phi(z|x)$ we work on estimating the logarithm of  $p_{\theta}(x)$ (see also \eqref{eq:prob_x} in Appendix B),

\begin{equation}
    \begin{split}
        \log p_\theta (x) &= \log \mathbb{E}_z\Big[p_{\theta}(x|z)\Big] 
        = \mathbb{E}_{z \sim q_\phi (z|x)} \Bigg[\log p_\theta(x) \Bigg]  \nonumber \\
        &= \mathbb{E}_z\Bigg[\log p_\theta (x|z)\Bigg] - \mathbb{E}_z\Bigg[\log \frac{q_\phi(z|x)}{p_\theta(z)}\Bigg] + \mathbb{E}_z\Bigg[\log \frac{q_\phi(z|x)}{p_\theta(z|x)} \Bigg]. 
    \label{eq:log4}
    \end{split}
\end{equation}

This however is a typical application for the Kullbag-Leibler (KL) divergence (see Appengix C or \cite{SK}). The Kullback-Leibler divergence is a semi-metric \cite{SK} and can provide an estimate of the distance between two measures $P$ and $Q$,
\[
D_{KL}(P||Q) = \mathbb{E}_{x\sim P}\Bigg[\log\frac{P(x)}{Q(x)}\Bigg].
\]
Using the KL divergence therefore we can express $\log p_\theta(x)$ as, 

\begin{align}
        \log p_\theta(x) = \mathbb{E}_{z \sim q_\phi (z|x)}\Big[\log p_\theta(x|z) \Big] - D_{KL}(q_\phi(z|x) || p_{\theta}(z)) +  D_{KL}(q_\phi(z|x) || p_\theta(z|x)).
    \label{eq:log5}
\end{align}
The first term above is possible to compute directly from the decoder and sampling with the reparameterization trick \cite{VAE}, which will be discussed later. The second term is a KL-divergence between a general and a normal Gaussian which has a closed form solution based on Property \ref{prop:KL_gauss} of Appendix C. Finally, the problematic last term can be shown to be greater than 0 \cite{Tiao} through Property \ref{prop:KL} in Appendix C. Therefore we can 
define the lower bound of \eqref{eq:log5} as,
\begin{align}
        L(\theta, \phi; x) = \mathbb{E}_{z\sim q_\phi (z|x)}\Big[\log p_\theta(x|z) \Big] -  D_{KL}(q_\phi(z|x) || p_{\theta}(z)).
        \label{eq:elbo}
\end{align}
Thus our task is to minimize the above with respect to the parameters $\theta$ and $\phi$,
\begin{equation}
\phi^*, \theta^* = \argmin_{\phi, \theta} L(\theta, \phi; x).
\label{eq:phi_theta_estimate}
\end{equation}
The only remaining issue is to resolve the expected value in \eqref{eq:elbo}. This can be accomplished with a reparameterization idea from 
\cite{VAE} provided below for completeness.

\subsection{The change of variables idea}

Functions such as (\ref{eq:elbo}) are not uncommon in optimization problems \cite{Helmholtz}. The novel idea, introduced in \cite{VAE}, which removes  difficulties of the expected value in 
\eqref{eq:elbo} is to introduce a suitable change of variables.

We express $z\sim q_\phi(z|x)$ through a deterministic transformation $g_\phi$ and a random variable $\epsilon \sim p(\epsilon)$ as 
$
z = g_\phi(x, \epsilon),
$
where $p(\epsilon)$ is a simple distribution which does not depend on $x$ or $\phi$. In our case we choose $p(\epsilon) = N(0,1)$ since we want the latent space to be Gaussian. This reduces $z$ to,
\begin{equation*}
z = g_\phi(x,\epsilon) = \mu_\phi(x) + \sigma_\phi(x) \odot \epsilon, \quad \epsilon \sim N(0, 1) 
\end{equation*}
We now define an auxiliary function $f(x, z) = \log p_\theta(x, z) - \log q_\phi(z|x)$ over the distribution $q_\phi(z|x)$ and write 
\eqref{eq:elbo} as an expectation of $f$. 
If we then substitute $z$ and compute the gradient with respect to $\phi$ we get,
\begin{align*}
    \nabla_\phi \mathbb{E}_{z \sim q_\phi(z|x)}[f(x,z)] = 
     \nabla_\phi \mathbb{E}_{g_\phi(x, \epsilon) \sim p(\epsilon)}[f(x, g_\phi(x, \epsilon))] 
    = 
    \mathbb{E}_{g_\phi(x, \epsilon) \sim p(\epsilon)}[\nabla_\phi f(x, g_\phi(x, \epsilon))].
\end{align*}
The expectation and the gradient commute which allows us to practically optimize the above using back propagation.


\subsection{The Conditional Latent Space Variational Autoencoder\label{sec:cvae}}


VAEs are designed to estimate the unknown probability distribution $p_\theta(x)$ for the given $x$ using a Gaussian in latent space. Instead we propose to use  information which is already available in the data in order to more accurately express the dataset $x$ through several Gaussians. We specify the Gaussians in latent space by conditioning on any one of the labels in the data set. For the MNIST data set of images for instance we condition on the number label associated with each image. 
Therefore for $N=10$ categories in that label in the data,
\begin{equation}
p_\theta(x) = \sum_{y=1}^N \int p_\theta(x|y,z)p_\theta(z|y)p_\theta(y)dz.
\label{eq:proposed}
\end{equation}
where we chose $p_\theta(y)$ to be a weighted uniform distribution since the numbers in the MNIST data are discrete constants and $p_\theta(z|y)$ is a set of $N$ Gaussians each with unknown mean and variance. Finally to understand $p_\theta(z|x,y)$ we apply Bayes rule,
\begin{equation}
p_\theta(z|x,y) = \frac{p_\theta(x|y,z) p_\theta(z|y) p_\theta(y)}{p_\theta(x)},
\end{equation}
where $p_\theta(z|x,y)$ can be estimated from the encoder $q_\phi(z|x,y)$. Note furthermore that $p_\theta(y)$ is known and gives $p_\theta(z|y)$. We can therefore write $p_\theta(x|y,z) = p_\theta(x|z)$.

Following arguments similar to constructing the loss function for VAE we now produce the following loss function for our CL-VAE,
\begin{equation}
    \begin{split}
        L(\theta, \phi; x) = \gamma \mathbb{E}_{z\sim q_\phi (z|x)}\Big[\log p_\theta(x|z) \Big] -  \beta D_{KL}(q_\phi(z|x,y) || p_{\theta}(z|y)).
    \label{eq:elbo2}
    \end{split}{}
\end{equation}
where $\gamma$ and $\beta$ are scalar weights that is going to be helpful in the next section. Since $y$ is discrete the KL divergence above can be written as,
\begin{equation*}
\begin{split}
    D_{KL}(q_\phi(z|x,y) || p_{\theta}(z|y)) =  \mathbb{E}_{z|y}\Bigg[\log\frac{q_\phi(z|x,y)}{p_\theta(z|y)}\Bigg] =  
    \sum_{y=1}^N p_\theta(z|y)\log\frac{q_\phi(z|x)}{p_\theta(z|y)}
\label{eq:KL2}
\end{split}{}
\end{equation*}
To produce a closed form solution for $D_{KL}(q_\phi(z|x,y) || p_{\theta}(z|y))$ in the non-normal Gaussian case we follow ideas from \cite{one-step} and let $q_\phi(z|x)$ be some Gaussian distribution that is conditioned on $x$,
	$$
	q_\phi(z|x) = \frac{\exp\Big\{-\frac{1}{2} \Big|\Big|\frac{z-\mu(x)}{\sigma(x)}\Big|\Big|^2 \Big\}}{\prod_{i=1}^d \sqrt{2\pi\sigma_i^2(x)}} 
	\label{prop:sol_dkl2}
	$$
	We now let $p_\theta(z|y)$ be Gaussian with mean $\mu_y$ and variance 1,
	\[
	p_\theta(z|y) = \frac{1}{(2\pi)^{d/2}} \exp\Big\{-\frac{1}{2}||z-\mu_y||^2 \Big\}
	\]
	and use in simplifying,
	\[
	\begin{split}
	    \log\frac{q_\phi(z|x)}{p_\theta(z|y)} =  -\frac{1}{2}\sum_{i=1}^d \log\sigma_i^2(x)  
	    - \frac{1}{2} \Bigg|\Bigg|\frac{z-\mu(x)}{\sigma(x)}\Bigg|\Bigg|^2 + \frac{1}{2} ||z-\mu_y||^2.
	\end{split}{}
	\]
We apply the same reparameterization idea as before with $z = g_\phi(x,\epsilon) = \mu_\phi(x) + \sigma_\phi(x) \odot \epsilon$ where $\epsilon \sim N(0, 1)$ which simplifies the expression above to, 
	\[
	\log\frac{q_\phi(z|x)}{p_\theta(z|y)} = -\frac{1}{2}\sum_{i=1}^d \log\sigma_i^2(x) + \frac{1}{2} ||z-\mu_y||^2.
	\]
Computation of $\sum_{y=1}^N p(z|y) \log \frac{q_\phi(z|x)}{p_\theta(z|y)}$ now reduces to a simple matrix operation.



\section{Clustering and Anomaly Detection Results}
\begin{figure}[H]
	\centering
	\includegraphics[width=.34\linewidth,,height=.28\linewidth]{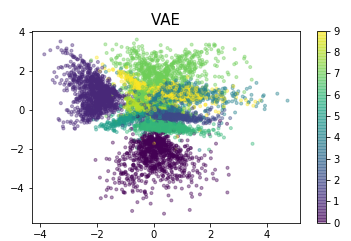}\hspace{-.3cm}
	\includegraphics[width=.34\linewidth,height=.27\linewidth]{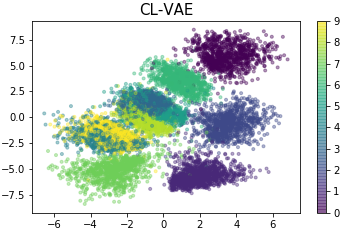}\hspace{-.3cm}
	\includegraphics[width=.34\linewidth,height=.27\linewidth]{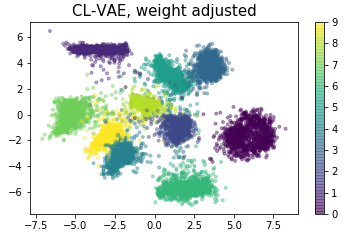}
	\caption{The 2D-latent space for the MNIST test data. Here the colors represent the 10 number classes for the VAE (left), the CL-VAE (middle) and a weight adjusted CL-VAE (right). The CL-VAE produces a clear latent space fitting and automatic separation between classes.}
	\label{fig:vanilla_vae}
\end{figure} 

Training a regular VAE on the MNIST dataset we are treated by a latent space representation where the classes accumulate around a single Gaussian. The model is trained on  minimizing the loss function \eqref{eq:elbo} with a single Gaussian normal prior. The result of this is clear in the top of Figure \ref{fig:vanilla_vae} as the points are all expanding outward from the origin where the normal Gaussian has its mean. As can be seen in that figure, there is some structure and class separation but clusters have inconsistent shapes and are overlapping heavily. Class overlap is natural since numbers can easily resemble each other and can be hard to tell apart for humans as well. However, class overlap will remain a problem when using the VAE. This is attributed to the fact that classes are trying to adapt their form to a single Gaussian prior.

We also train the CL-VAE on the same dataset and present the results in the middle of Figure \ref{fig:vanilla_vae}. It is immediately evident, at least visually, that the class separation has now improved compared with that from a regular VAE. The class overlap is still visible since numbers due to their inherent shape resemble each other in the MNIST data. In that respect it is not surprising to see that the distribution for numbers $6$ and $0$ are close to each other and the clusters forming 4 and 9 are completely overlapping. We also see that $4$ and $9$ seem to be completely overlapping and 3, 5 and 7 are partially overlapping. This is not going to be helpful for clustering.

Finally, we also show a weight adjusted CL-VAE to the right in Figure \ref{fig:vanilla_vae}. Here the reconstruction error is weighted down s.t. $\gamma = 0.3$ and $\beta=1.0$. This means that more focus is put on the KL-divergence term in training. Visually, we can see that $4$ and $9$ no longer overlap and no other class is completely overlapping with any other. 

\subsection{Clustering\label{clustering}}
We now present and analyze a number of clustering results comparing our proposed CL-VAE and a regular VAE. During our analysis we use two different clustering methods: the $k$-means and the expectation maximization (EM). A visual inspection of the results is always helpful although we also  measure the V-score in order to better assert the success of each algorithm.



\begin{figure}[h]
    \centering
    \includegraphics[width=.34\linewidth,height=.27\linewidth]{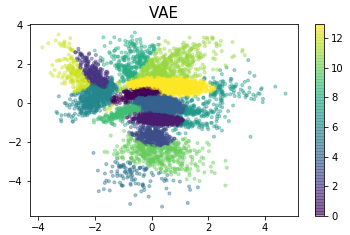}\hspace{-.23cm}
    \includegraphics[width=.34\linewidth,height=.27\linewidth]{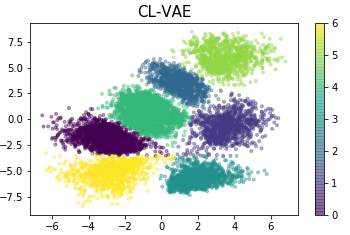}\hspace{-.32cm}
    \includegraphics[width=.34\linewidth,height=.27\linewidth]{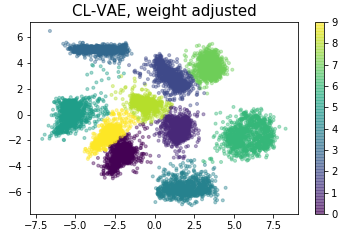}
    \caption{The results of performing EM-clustering on the three different latent spaces of Figure \ref{fig:vanilla_vae} using the optimal number of clusters from Table \ref{tab:clustering}. Comparing this Figure with Figure \ref{fig:vanilla_vae} we can see which numbers ended up in each of the clusters. Note that the numbers 4 and 9 ended up in the same cluster for the standard CL-VAE.}
    \label{fig:clustering}
\end{figure}{}

Based on the latent space representations from VAE and CL-VAE we compute their respective V-scores in order to find the optimal number of clusters for each. For the VAE, the V-score is quite bad for both the EM and $k$-means clustering algorithms as can be seen in Table \ref{tab:clustering}. Another problem is that it reaches its maximum V-score using 14 clusters, more than the true amount of numbers in the dataset. This however is not that strange since the clusters based on the VAE latent space shown in Figure \ref{fig:vanilla_vae} are not clearly separated.

For the CL-VAE we get a higher V-score for both the EM and $k$-means algorithms but the clusters are now too few instead. Inspecting Figure \ref{fig:clustering} and comparing it with Figure \ref{fig:vanilla_vae} we see that the numbers 4 and 9 ended up in the same cluster. The same is true for the numbers 3, 5 and 7 as well. 

Finally, considering the weight adjusted CL-VAE we find that we get 10 unique clusters and again, a much higher V-score for both EM and $k$-means. Using this setup you can therefore actually cluster the numbers quite successfully. As a result we will use the weight adjusted CL-VAE to identify anomalies in the next section. 

\begin{table}[H]
    \centering
    \begin{tabular}{|c|c|c|c|c|}
    \hline
        Algorithm & \multicolumn{2}{c|}{EM} & \multicolumn{2}{c|}{$k$-means} \\
        \hline
        \hline 
        VAE & 0.6313 & 14 & 0.5572 & 14 \\
        \hline
        \hline
        CL-VAE & 0.8132 & 7 & 0.8088 & 7 \\
        \hline
        $\gamma=0.3$ & \textbf{0.9248} & \textbf{10} & 0.9233 & 10 \\ \hline
    \end{tabular}
    \caption{The results of clustering for each of the 3 different latent spaces. The left column under type of clustering algorithm is the optimal V-score and the right column is the optimal number of clusters. As is evident, the CL-VAE with the reconstruction error weighted down performs the best (boldface).}
    \label{tab:clustering}
\end{table}{}

\subsection{Anomaly detection\label{sec:anomaly} in MNIST}
In this section we present and compare unsupervised anomaly detection capabilities based on the latent space of the proposed CL-VAE versus that of the regular VAE. We perform these comparisons under the context of three of the most widely used methods for anomaly detection. We also note that neither of our datasets include information or labels as to whether a given data point is an anomaly or not. 

Some of the algorithms applied here to identify anomalies include hyper-parameters which must be tuned. We have therefore undertaken this task and in all results presented we have already established the best set of such hyper-parameters for those methods. 

\begin{table}[h]
    \centering
    \begin{tabular}{|c|c|c|}
        \hline
        LS/Algo. & EM$_s$ \hspace{.1cm}($10^{-3}$) & MV$_s$ \\
        \hline
        \hline
        \begin{tabular}{c}
            \textbf{VAE} \\
            \hline
            IF \\
            LOF \\
            OCSVM
        \end{tabular}{}
        &
        \begin{tabular}{c}
            \\
             $2.532$ \\
             2.014  \\
             \textbf{2.533}
        \end{tabular}{}
        &
        \begin{tabular}{c}
            \\
             2.465  \\
             3.136  \\
             \textbf{2.442}
        \end{tabular}{}
        \\
        \hline
        \hline
        \begin{tabular}{c}
            \textbf{CL-VAE} \\
            \hline
            IF \\
            LOF \\
            OCSVM
        \end{tabular}{}
        &
        \begin{tabular}{c}
            \\
             0.884  \\
             0.841  \\
             \textbf{0.926}
        \end{tabular}{}
        &
        \begin{tabular}{c}
            \\
             7.065  \\
             7.236  \\
             \textbf{6.754}
        \end{tabular}{} \\
        \hline
    \end{tabular}
    \caption{Comparisons of excess mass (EM$_s$) and mass volume (MV$_s$) scores for three different anomaly detection methods based on either the VAE or the CL VAE latent space of MNIST. The results indicate that the one-class support vector machine (OCSVM) performs best for both latent spaces. However, during further investigation OCSVM was ruled out as it classifies about half of the data as anomalous. }
    \label{tab:em-mv-summary}
\end{table}{}

Remembering that we should maximize the EM$_s$ score and minimize the MV$_s$ score, the results presented in Table \ref{tab:em-mv-summary} clearly indicate that the VAE produces better EM-MV scores than the CL-VAE. This is rather unexpected and something we'll discuss further. It's also clear that the OCSVM provides the best EM-MV scores for both types of latent spaces. Investigating a little further, we find that the OCSVM classifies more than half of the test data as anomalous. This obviously deems it rather useless.

Ruling out the OCSVM, the Isolation Forest algorithm performs the best on both latent spaces. Closer investigation of the  anomaly classification mechanism however reveals that the IF algorithm classifies the center of the prior distribution for the VAE as normal and the edges of the distribution as anomalous. This means that, while the number of elements in each class are approximately the same, the numbers 0, 1 and 7 are over-represented among the anomalies while numbers 6, 9, 2, 8 and 5 are under-represented. Taking a look at the VAE in Figure \ref{fig:vanilla_vae} again the reason for this becomes clear as there is plenty of class overlap in the center of the distribution for the VAE where 6, 9, 2, 8 and 5 are all put in the middle. At the same time, 0, 1 and 7 occupy the fringes of the distribution, leading to the over-representation in the number of anomalies. 

This behavior is not as prevalent in the  LOF algorithm which actually finds some anomalies between clusters, leading to a more well-balanced distribution of anomalies among the different classes. So even though the LOF performed the worst on the EM-MV score, it seems to be the most reasonable classifier of anomalies in this case.

\begin{figure}[H]
    \centering
    \includegraphics[width=0.6\linewidth]{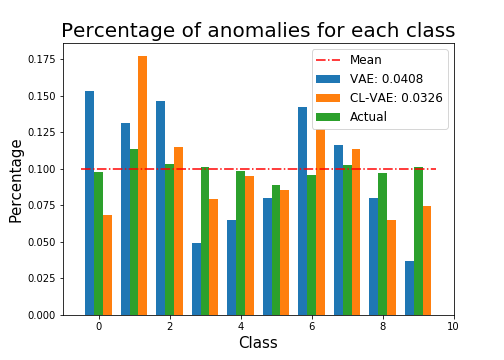}
    \caption{A bar chart comparing the anomalies (in percent) found within each class when performing anomaly detection on the two different latent spaces. The CL-VAE gets closer to the actual distribution and therefore gets a lower RMSE, which is the number displayed in the box next to each algorithm.}
    \label{fig:percentage}
\end{figure}{}

Now considering the weight adjusted CL-VAE, we find that again disregarding the OCSVM, Isolation Forest gets the best EM-MV scores. However, the same over- and under-representation problems apply here. It seems like it weighs the regions with classes that are slightly overlapping as 'more normal' while classes that are clearly distinct are considered 'more anomalous'. For instance, it finds 144 anomalies for the number 6's while only 14 anomalies for the number 8. Once again however LOF manages to provide us with a much more even distribution of anomalies as seen in Figure \ref{fig:percentage}. 

\begin{figure}[H]
    \centering
    \includegraphics[width=.48\linewidth]{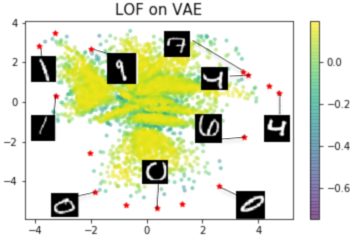}
    \includegraphics[width=.48\linewidth]{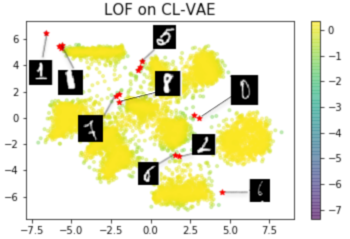}
    \caption{The results of running anomaly detection using the LOF algorithm on the two different latent spaces. In this case, the 15 most anomalous points are highlighted with a red star and some are filled in with their corresponding image in MNIST. As is clear, the VAE tends to overemphasize certain numbers in anomaly detection as was described earlier, that tend to end up on the fringes of the prior distribution. While the CL-VAE finds it's most uncommon anomalies between clusters that tend to be more unclear of what class they actually belong to.}
    \label{fig:lof_anomalies}
\end{figure}

Overall however we see that in general VAE produces better EM-MV scores than CL-VAE. This could be because the VAE latent space has a few points that end up far out on the tail of the distribution and therefore the algorithms have an easier time distinguishing normal points from anomalous. That being said, these points are mostly consisting of a few classes, leading to a more unbalanced set of anomalies which is clearly wrong. That is where the CL-VAE has an advantage. To visualize this, we have plotted the percentages of anomalies for each class in Figure \ref{fig:percentage}. What is clear from this plot is that both autoencoders over-estimate and under-estimate certain classes. But overall, the CL-VAE get's a lower RMSE from the actual distribution of the dataset at 0.0326 compared with 0.0408 for the VAE.

To summarize our findings in this section we found that the VAE separates the data in a more extreme way than CL-VAE while at the same time overestimating anomalies for some classes at the edges of its single-blob type latent space. This higher degree of separation seems to be why the VAE achieves better EM-MV scores. The CL-VAE produces a more balanced set of anomalies in the dataset which stem for all of the classes in the data. Considering Figure \ref{fig:lof_anomalies}, the most extreme anomalies seem more difficult to categorize than those anomalies identifies by the VAE. 

\subsection{Anomaly Detection in Trading Data}
The dataset of trades from \cite{erno} contains about 100 000 rows and has columns describing what type of trade occurred. These include margin, nominal value, currency, type of instrument, counter party, portfolio and importantly \textit{trader name}. By conditioning on what trader did what trade, we can try and create similar clusters to Figure \ref{fig:vanilla_vae}. Instead of variations of handwritten numbers, the clusters describe the trading behavior of each trader in the dataset. Two overlapping traders for instance suggest that they traded in a similar fashion, meaning we have found a trading category. For futher details related to the data we refer to the thesis work in \cite{erno}.
\begin{table}[h]
    \centering
    \begin{tabular}{|c|c|c|}
        \hline
        LS/Algo. & EM$_s$ \hspace{.1cm}($e^{-3}$) & MV$_s$ \\
        \hline
        \hline
        \begin{tabular}{c}
            \textbf{VAE} \\
            \hline
            IF \\
            LOF \\
            OCSVM
        \end{tabular}{}
        &
        \begin{tabular}{c}
            \\
             4.648  \\
             3.587 \\
             \textbf{5.903}
        \end{tabular}{}
        &
        \begin{tabular}{c}
            \\
             1.362  \\
             1.736 \\
            \textbf{1.053}
        \end{tabular}{}
        \\
        \hline
        \hline
        \begin{tabular}{c}
            \textbf{CL-VAE} \\
            \hline
            IF \\
            LOF \\
            OCSVM
        \end{tabular}{}
        &
        \begin{tabular}{c}
            \\
             \textbf{16.33}  \\
             13.48  \\
             12.43
        \end{tabular}{}
        &
        \begin{tabular}{c}
            \\
             \textbf{0.5168}  \\
             0.5509  \\
             0.567
        \end{tabular}{} \\
        \hline
    \end{tabular}
    \caption{Comparisons of excess mass (EM$_s$) and mass volume (MV$_s$) scores for three different anomaly detection methods based on either the VAE or the CL-VAE latent space using the trading data. The results indicate that OCSVM performs best for VAE while Isolation Forest performs best for the CL-VAE. We note that all algorithms performed better on the latent space of the CL-VAE giving merit to our proposed method.}
    \label{tab:em-mv-summary-shb}
\end{table}{}

\begin{figure}[H]
    \centering
    \includegraphics[width=0.45\linewidth,height=.31\linewidth]{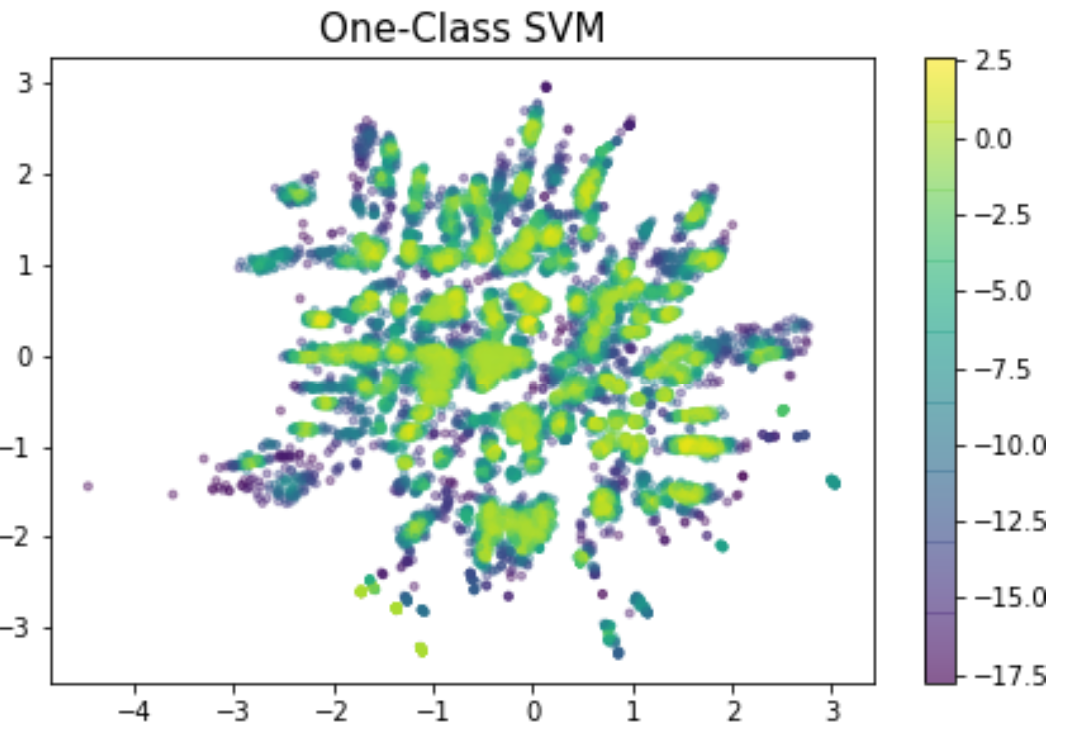} 
    \includegraphics[width=0.45\linewidth,height=.32\linewidth]{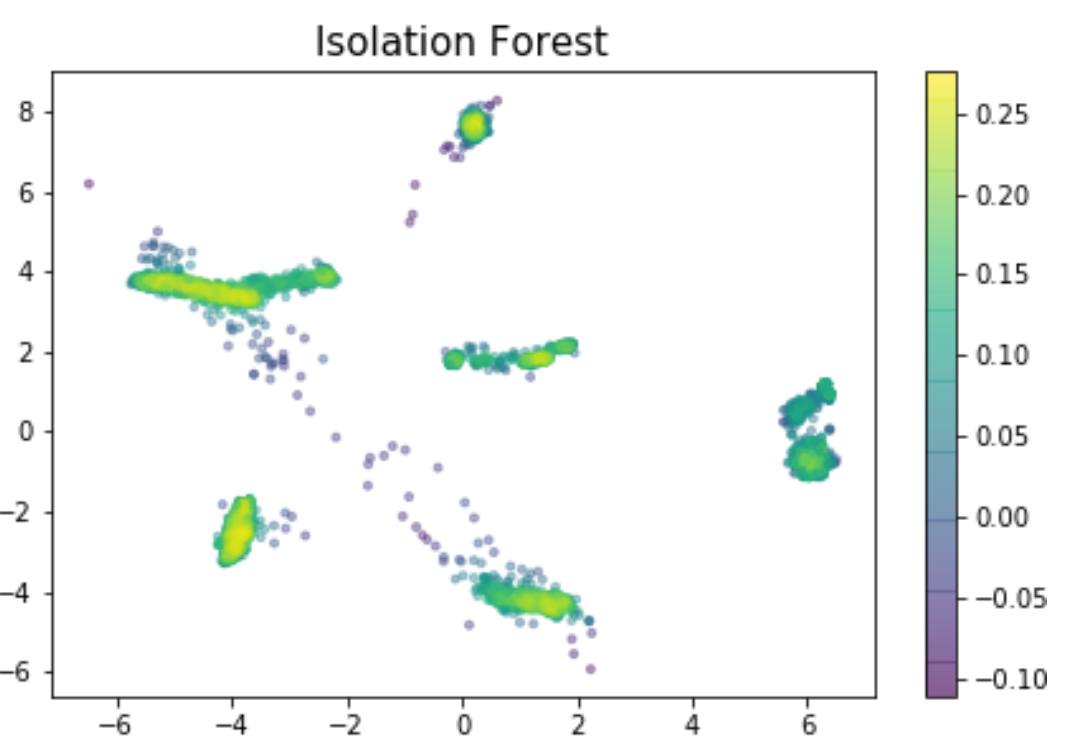}
    \caption{Comparing the two different latent spaces of the VAE (left) and the CL-VAE (right) of the trading data. The color coding corresponds to the best performing anomaly detection algorithm on each latent space.}
    \label{fig:anomaly-detect-shb}
\end{figure}

As can be seen in Table \ref{tab:em-mv-summary-shb} we see that, on this dataset, the CL-VAE latent space outperforms the VAE latent space on the EM-MV measure for all algorithms tried. One big difference here is that the clusters that was formed with CL-VAE are placed much further apart than the clusters in MNIST as can be seen in Figure \ref{fig:anomaly-detect-shb}. This is likely contributing to the higher EM-MV score as it seems easier for the anomaly detection algorithms to contrast normal from anomalous points in such a latent space.

\subsection{Using Misclassification for Anomaly Detection}
We now explore an alternative way toward anomaly detection. If a cluster is found in latent space where one most frequent class can be identified, then the points inside this cluster which do not belong to the most frequent class can be considered anomalies. To identify these points however can be tricky. These are points that in some sense behave more like the predicted class than its own class in the conditioned space, meaning that they should be classified as anomalies.  

We will employ the V-score to assist us in scoring and detecting these anomalies. As we showed in Table \ref{tab:clustering}, the CL-VAE has a much better V-score than the VAE. In the case of the MNIST data this means that numbers which are classified to belong to a cluster will be more likely to have the same label (homogeneity) and all of the members of that class are more likely to be assigned to the same cluster (completeness).

\begin{figure}[H]
    \centering
    \includegraphics[width=0.15\linewidth]{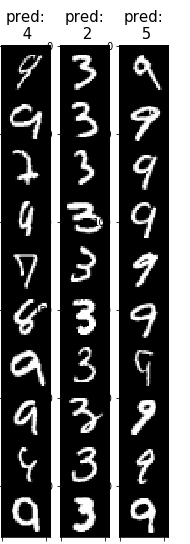}
    \includegraphics[width=0.15\linewidth]{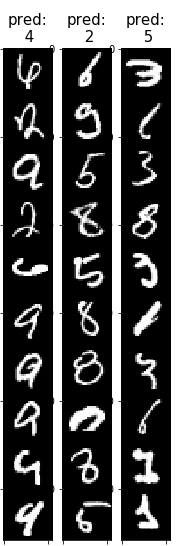}
    \caption{A sample of misclassified numbers for each latent space. For the VAE many normal looking numbers were given the wrong label. Note also that mostly 3s, 7s and 9s were misclassified. Meanwhile, the misclassifications in the CL-VAE look more unusual and have a more even mix of numbers.}
    \label{fig:misclassified}
\end{figure}

We provide a representative sample of anomaly misclassification based on both the VAE and CL-VAE latent spaces in Figure \ref{fig:misclassified}.
As can be seen in that figure some of the handwritten numbers which were classified as 'anomalies' using the VAE latent space are rather easy to identify. Some would probably be considered anomalies but certainly not most. This is not the case for the CL-VAE where many of the found anomalies are in fact unintelligible. 


\begin{figure}[H]
    \centering
    \includegraphics[width=.5\linewidth]{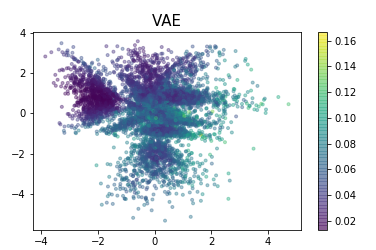}
    \includegraphics[width=.48\linewidth]{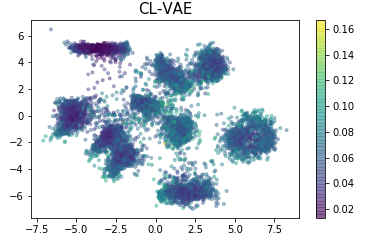}
    \caption{The latent space of the VAE (top) and the CL-VAE (bottom) colored with the RMSE of the numbers with respect to their means.}
    \label{fig:mean_deviation}
\end{figure}

One of the big points about the CL-VAE is that points that end up on the tails of their respective prior distributions should be 'more anomalous'. There's no such guarantee with the VAE, because it only has one prior distribution. Instead, giving each class their own prior distribution should mean that the classes don't have to fight over placement in the latent space. To illustrate this further, we coloured the latent spaces according to the average deviation for each point compared with the respective mean of each class in Figure \ref{fig:mean_deviation}. 

What this means is that points that are closer to the mean of each class also end up closer to their respective means in the latent space of the CL-VAE. If we compare it with the VAE, we see that the middle of the prior distribution actually tends to have a higher error than on some pockets on the tail of the distribution. Ultimately, what this means is that the latent space of the VAE can't be interpreted as samples from a Gaussian Normal distribution. Because there is no clear structure of where the points should end up based on their 'degree of abnormality'. This is in contrast with the CL-VAE, where the points can be interpreted as being sampled from a GMM prior. Consequently, points that either end up on the tail of their respective prior or end up in the wrong cluster, can in fact be called anomalies.

\section{Discussion\label{discussion}}



In this paper we explore classification of labeled data by conditioning in latent space. Conditioning allows us to use information within the data to improve clustering. We subsequently use these clusters to identify anomalies in the data. To showcase our findings we used the MNIST data set as well as actual trades from traders in a Swedish bank.

Our overall strategy for classification and anomaly detection involves a number of methodologies. We begin with automatic formation of clusters in latent space. These clusters are found and shaped according to optimal recommendations from our CL-VAE. This improved clustering description allows us to apply a number of methods for detecting anomalies. Anomaly detection for us relies on detecting outliers in the latent space of the autoencoder. We compare Isolation Forest, one-class SVM and LOF in order to identify the outliers for each of those clusters and find the optimal model for each case.

We also note that neither of the datasets used was labeled in terms of whether a given point is an anomaly or not. 
This led us to propose methodologies which while unsupervised can indicate whether a given trade is typical or anomalous.
The EM-MV measure is able to measure the performance of anomaly detection of the different models. However, it does not capture the full story as we saw in Section \ref{sec:anomaly}.
Furthermore the proposed methodology works with data of categorical nature (non-continuous) in order to find meaningful latent representations - both of which are not possible for classic PCA methods \cite{PCA}.

The proposed CL-VAE makes the latent space easier to understand while clustering data and performing anomaly detection. The question of how much of the proposed methodology is truly unsupervised should also be addressed. Clearly we are using labeling in the data to help us in the \emph{initial} classification. So this is not an entirely unsupervised method. The data is forced into Gaussians based on that information. For the MNIST data for instance that corresponds to 10 clusters. However the methodology proposed is not completely supervised either since after the conditioning is performed the method performs all actions in an unsupervised way based on a newly formed latent space \cite{Geron, Chollet}.

We have shown that a weight adjusted CL-VAE can be a successful pre-processing procedure for clustering and eventual anomaly detection. It strongly outperformed the VAE on the clustering task. While the VAE gets good EM-MV scores on the MNIST dataset, it does so while overestimating some classes and underestimating others. This problem is decreased when using the CL-VAE. Also, the misclassified numbers in the CL-VAE do look much more unintelligible than the ones found for the VAE. Meanwhile, using the trading dataset, the CL-VAE obtained better EM-MV scores than the VAE.

The proposed CL-VAE attempts to make the latent space more understandable and suitable for analysis with established methods. It seems to succeed in this regard, as it both divides the latent space up in accordance with class labels and ensures that points that are improbable do in fact end up on the tail of their respective prior distributions or in other clusters. 

It would be interesting to study the generative aspects of the CL-VAE as well in the future. In particular, it would be interesting to compare it with a CVAE, as they both have the functionality of generating samples from a given class, something that the VAE does not. 

\newpage
\section*{Appendix A. Background on Autoencoders}

An autoencoder is a neural network capable of performing unsupervised dimensionality reduction. As a result it is able to discover a lower-level representation of a higher dimensional data space. 

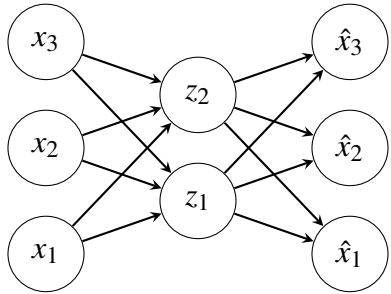
\begin{figure}[H]
	\centering
	\begin{tikzpicture}[node distance=2cm]
	\node (output 1) [in_node, yshift=-40] {$\hat{x}_1$};
	\node (output 2) [in_node] {$\hat{x}_2$};
	\node (output 3) [in_node, yshift=40] {$\hat{x}_3$};
	\node (input 1) [in_node, left of=output 1, xshift=-2cm] {$x_1$};
	\node (input 2) [in_node, left of=output 2, xshift=-2cm] {$x_2$};
	\node (input 3) [in_node, left of=output 3, xshift=-2cm] {$x_3$};
	\node (layer 1) [in_node, left of=output 2, yshift=-20] {$z_1$};
	\node (layer 2) [in_node, left of=output 2, yshift=20] {$z_2$};
	\draw [arrow] (input 1) -- (layer 1);
	\draw [arrow] (input 2) -- (layer 1);
	\draw [arrow] (input 3) -- (layer 1);    
	\draw [arrow] (input 1) -- (layer 2);
	\draw [arrow] (input 2) -- (layer 2);
	\draw [arrow] (input 3) -- (layer 2);
	\draw [arrow] (layer 1) -- (output 1);
	\draw [arrow] (layer 1) -- (output 2);
	\draw [arrow] (layer 1) -- (output 3);    
	\draw [arrow] (layer 2) -- (output 1);
	\draw [arrow] (layer 2) -- (output 2);
	\draw [arrow] (layer 2) -- (output 3);
	\end{tikzpicture}
	\caption{An idealization of an autoencoder. Here $x_i$ is the input, $z_i$ the hidden representation with 2 latent dimensions and $\hat{x}_i$ the reconstruction of the input $x_i$.}
	\label{fig:autoencoder}
\end{figure}

An autoencoder is typically constructed of two neural networks: an encoder and a decoder. An idealized autonencoder representation is given in Figure \ref{fig:autoencoder} where we reduce a 3 dimensional input to a 2 dimensional latent space.

In general autoencoder applications both the encoder and the decoder networks are densely connected. 

\begin{definition} \label{def:autoencoder}
	Autoencoder. Given an input $x \in \mathbb{R}^d$ we assume that there is a mapping $E$ to $z \in \mathbb{R}^p$ s.t. $E : x \rightarrow z$. This mapping is called the encoder and can be defined with an activation function $\sigma$, a weight matrix $W$ and a bias term $b$ as,
	\[
	z = \sigma ( W x + b).
	\]
	Conversely, we assume that there is a mapping $D$ s.t. $D : z \rightarrow x$ which is called the decoder and can be defined in a similar way to the encoder with the corresponding terms $\hat{\sigma}$, $\hat{W}$ and $\hat{b}$ as,
	\[
	\hat{x} = \hat{\sigma} (\hat{W} z + \hat{b}).
	\]
\end{definition}

\begin{figure}[H]
	\centering
	\begin{tikzpicture}[node distance=2cm]
	\node (output) [in_node] {$\hat{x}$};
	\node (decoder) [node, left of=output, xshift=-20] {Decoder};
	\node (z) [in_node, left of=decoder, xshift=-20] {$z$};
	\node (encoder) [node, left of=z, xshift=-20] {Encoder};
	\node (input) [in_node, left of=encoder, xshift=-20] {$x$};
	\draw [arrow] (input) -- (encoder);
	\draw [arrow] (encoder) -- (z);    
	\draw [arrow] (z) -- (decoder);
	\draw [arrow] (decoder) -- (output);
	\end{tikzpicture}
	\caption{A flow chart of the autoencoder. Here the white circles represent input or output data and the grey rectangles represent neural networks.}
	\label{fig:my_label}
\end{figure}
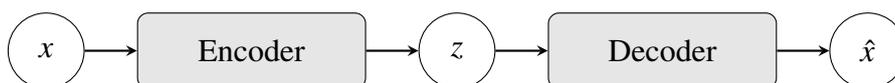


According to this definition therefore the autoencoder's job is to estimate a non-linear transformation from $x$ to $z$ and its inverse from $z$ to $\hat{x}$. In order for this autoencoder to compute these transformations it minimizes the reconstruction error of $x$,
\begin{align*}
\begin{split}
L(x, \hat{x}) = ||x - \hat{x}||^2 
=  ||x-\hat{\sigma} (\hat{W} z + \hat{b})||^2 
= ||x-\hat{\sigma} (\hat{W} (\sigma ( W x + b)) + \hat{b})||^2.
\end{split}
\end{align*}
In general however autoencoders produce latent spaces which are not useful for analysis \cite{Chollet}. This is one of the main reasons that variational autoencoders have gained such popularity.

\section*{Appendix B. Variational autoencoders and Bayes rule}

Variational autoencoders (VAEs) are able to discover the distributions responsible for the provided data. A VAE therefore solves the problem of probability density estimation and is a true generative model. This practically means that you can generate new samples from an unknown distribution \cite{Stanford}. Applications in image processing for example use VAEs to generate new images which retain some of the main features of the original data set \cite{TrainVAE}.


If we have some set of locally observed variables $x$ and we assume that they follow some unknown stochastic process $X$ that we want to sample from, we can use some prior $z$ that we assume to be Gaussian. By then taking the expectation of the conditional distribution of $x$ given $z$, under $z$, we get the distribution for $x$ from,
\begin{equation}
p_{\theta}(x) = \mathbb{E}_z\Big[p_{\theta}(x|z)\Big] = \int p_{\theta}(x|z) p_{\theta}(z) dz.
\label{eq:prob_x}
\end{equation}
Note however that the integral above is intractable \cite{VAE}. This is where Bayes rule can help.

We assume that the density functions  $p(x)$ and $p(y)$ for stochastic variables x and y are known. If the conditional density function $p(x|y)$ is given then the conditional density function $p(y|x)$ can be computed from,
\begin{equation}
\label{def:bayes}
	p(y|x) = \frac{p( x| y) p(y)}{p(x)}.
\end{equation}

Bayer rule will be useful in terms of computing the posterior $p_{\theta}(z|x)$,
\begin{equation}
p_{\theta}(z|x) = \frac{p_{\theta}(x|z) p_\theta(z)}{p_\theta(x)}.
\label{eq:posterior}
\end{equation}
However $p_\theta(x)$ above is not possible to compute in general. Instead we estimate the posterior distribution $p_{\theta}(z|x)$ using some other distribution  $q_\phi(z|x)$ where $\phi$ contains estimates of the model parameters. 

\section*{Appendix C. Kullbag-Leibler divergence}

The KL-divergence has a number of imporant properties which we outline below. First we provide some definitions from information theory. 

Given two probability distributions $Q$ and $P$ the entropy of $P$ is defined from,
\begin{equation}
	H(P) = \mathbb{E}_{x\sim P}[-\log P(x)],
	\label{def:entropy}
\end{equation}
	and the cross-entropy of $Q$ and $P$ is given by,
\begin{equation}
	H(P, Q) = \mathbb{E}_{x\sim P}[-\log Q(x)].
\end{equation}

Given two probability distributions $Q$ and $P$ we define the Kullback-Leibler divergence by taking the cross-entropy minus the entropy,
\begin{align}
\label{def:KL}
\begin{split}
    D_{KL}(P||Q) = H(P, Q) - H(P) = 
    \mathbb{E}_{x\sim P}\Bigg[\log\frac{P(x)}{Q(x)}\Bigg].
\end{split}
\end{align}
The Kullback-Leibler divergence measures how well $Q$ approximates $P$.

\begin{property} \label{prop:KL}
	Properties of KL-divergence \cite{der}.
	\begin{enumerate}
		\item if $P=Q$ then $D_{KL}(P||Q) = 0$,
		\item if $P\neq Q$ then $D_{KL}(P||Q) > 0$.
	\end{enumerate}
\end{property}

\begin{property} \label{prop:KL_gauss}
	Solution to $D_{KL}(q(x)||p(x))$ in the normal Gaussian case \cite{VAE}. Let's assume that x is some random variable in  $q(x)\sim N(\mu, \sigma)$ and $p(x) \sim N(0, I)$ 
	\begin{align*}
	D_{KL}(q(x)||p(x
	)) = \mathbb{E}_{z \sim q_\phi(z)}\Bigg[\log\frac{q(x)}{p(x)}\Bigg] =   =\int \log \frac{q(x)}{p(x)} q(x) dx = 
	= \int (\log q(x) - \log p(x)) q(x) dx.
	\end{align*}
	This can now be evaluated as two different integrals. The first being
	\begin{align*}
	\int q(x) \log q(x) dx = 
	= \int N(x; \mu, \sigma^2)\log N(x; \mu, \sigma^2)dx = 
	= \frac{J}{2}\log(2\pi) + \frac{1}{2} \sum_{j=1}^J(1 + \log \sigma_j^2),
	\end{align*}
	and the second,
	\begin{align*}
	\int q(x)\log p(x)dx = 
	= \int N(x; \mu, \sigma^2) \log N(x; 0, I) dx =
	= \frac{J}{2} \log(2\pi) + \frac{1}{2} \sum_{j=1}^J (\mu_j^2 + \sigma_j^2),
	\end{align*}
	giving us the closed form solution,
	\begin{align*}
	D_{KL}(q(x)||p(x)) = 
	= -\frac{1}{2}\sum_{j=1}^J 1 + \log \sigma_j^2 - \mu_j^2 - \sigma_j^2.
	\end{align*}
\end{property} 


\end{document}